\documentclass[sigconf]{acmart}

\usepackage{booktabs} % For formal tables
\usepackage{mathtools}

\DeclarePairedDelimiter\floor{\lfloor}{\rfloor}

\setcopyright{none}
\acmYear{2021}
\copyrightyear{2021}

\begin{document}


%\title{Reinforcement Learning with Rare Rewards: \\ Learning to Cooperate with Partner Choice}
%\title{Reinforcement Learning with Rare Rewards: \\ Cooperation, Partner Choice and Biological Market}
%\title{Reinforcement Learning with Rare Rewards: \\ Learning to Cooperate with Partner Choice}
%\title{Reinforcement Learning with Rare Rewards: \\ Learning Partner Choice in Cooperation}
%\title{Reinforcement Learning with Rare Rewards: \\ Learning how to choose the right cooperative partner}
%\title{Reinforcement Learning with Rare Rewards: \\ Choosing the Right Partner to Cooperate with}
% RQ ==> (1) ce sont les évènements sont rares (la reward elle n'arrive qu'une fois) (2) policy search 
%\title{Policy Search with Rare Significant Events: \\ Choosing the Right Partner to Cooperate with}
\title{Policy Search with Rare Significant Events: \\ Choosing the Right Partner to Cooperate with}


% [ANONYMOUS]% [ANONYMOUS]% [ANONYMOUS]% [ANONYMOUS]% [ANONYMOUS]
% CAM-READY:
% CAM-READY: remettre: 
% CAM-READY:    (1) les Supmat
% CAM-READY:    le reste a été fait le 2021-02-05 après soumission GECCO
% CAM-READY:
% CAM-READY:
% CAM-READY:
% CAM-READY:
% CAM-READY:
% CAM-READY:
% CAM-READY:
% CAM-READY:
% CAM-READY:
% CAM-READY:
% CAM-READY:
% CAM-READY:
% CAM-READY:
% CAM-READY:
% CAM-READY:
% CAM-READY:
% CAM-READY:
% CAM-READY:
% CAM-READY:
% CAM-READY:
% CAM-READY:
% CAM-READY:
% CAM-READY:
% CAM-READY:
% CAM-READY:
% CAM-READY:
% CAM-READY:
% CAM-READY:

\author{Paul Ecoffet}
\affiliation{%
  \institution{Institut des Systèmes Intelligents et de Robotique, Sorbonne Université}
  \streetaddress{4 place Jussieu}
  \city{Paris}
  \country{France}
  \postcode{75005}
}
\email{paul.ecoffet@sorbonne-universite.fr}
\orcid{0000-0002-7394-6134}
\author{Nicolas Fontbonne}
\affiliation{%
  \institution{Institut des Systèmes Intelligents et de Robotique, Sorbonne Université}
  \streetaddress{4 place Jussieu}
  \city{Paris}
  \country{France}
  \postcode{75005}
}
\email{nicolas.fontbonne@sorbonne-universite.fr}
\author{Jean-Baptiste André}
\affiliation{%
  \institution{Institut Jean Nicod, Département d'Études Cognitives, École Normale Supérieure}
  \streetaddress{29 rue d’Ulm}
  \city{Paris}
  \country{France}
  \postcode{75005}
}
\email{jeanbaptisteandre@gmail.com}
\author{Nicolas Bredeche}
\affiliation{%
  \institution{Institut des Systèmes Intelligents et de Robotique, Sorbonne Université}
  \streetaddress{4 place Jussieu}
  \city{Paris}
  \country{France}
  \postcode{75005}
}
\email{nicolas.bredeche@sorbonne-universite.fr}


\begin{abstract}

This paper focuses on a class of reinforcement learning problems where significant events are rare and limited to a single positive reward per episode. 
A typical example is that of an agent who has to choose a partner to cooperate with, while a large number of partners are simply \textit{not} interested in cooperating, regardless of what the agent has to offer.
%, in order to accomplish a task together.
%The optimal policy implies that the agent is selective enough when choosing its partner \textit{and} is ready to invest enough energy for its partner to accept to cooperate.
%While the conditions for optimal partner choice are well known in the literature of theoretical biology, 
%However, learning how to choose the right partner can be challenging if the agent faces a large number of partners which simply are \textit{not} interested to cooperate.
We address this problem in a continuous state and action space with two different kinds of search methods: a gradient policy search method and a direct policy search method using an evolution strategy.
We show that when significant events are rare, gradient information is also scarce, making it difficult for policy gradient search methods to find an optimal policy, with or without a deep neural architecture. On the other hand, we show that direct policy search methods are invariant to the rarity of significant events, which is yet another confirmation of the unique role evolutionary algorithms has to play as a reinforcement learning method.

%, and produce optimal policies and converge faster.

%, both in performance and convergence speed. 

\end{abstract}



% From https://dl.acm.org/ccs#


\maketitle

\section{Introduction}

We consider a particular class of reinforcement learning problems where only rare events can result in non-zero rewards \textit{and} when the agent can experience at most one positive reward in a limited time. This problem is closely related to the problem of learning with \textbf{rare significant events} in reinforcement learning~\citep{bhatnagar2006, frank2008, ciosek2017}, where rare events can significantly affect performance (e.g. in network and communication systems or control problems where failure can be catastrophic). In this paper, we consider that while significant events occur independently of the agent's actions, the agent's policy determines if a positive reward should be obtained when such an event occurs. Significant events are thus defined as unique opportunities to obtain a positive reward and stop the game. Each opportunity can either be seized for an immediate reward or ignored if the agent hopes to get a better reward in the future.

We address this problem in the context of an \textbf{independent, on-line and on-policy episodic learning task} with continuous state and action spaces. The practical application addressed in this paper is that of an agent learning to choose a partner for a task that requires cooperation (e.g., predators hunting a large prey or individuals selecting a lifelong mate). The agent can choose to cooperate or not with a potential partner, based on the effort this partner is willing to invest in the cooperation. 
%whether the partner itself is a learning agent or not. 
At the same time, the agent must invest enough so that its partner also accepts to cooperate. In this setup, the agent may face partners willing to invest various amount of energy in cooperation (i.e., a possibly significant event), or even refuse to cooperate whatever the agent is ready to invest (i.e. a non-significant event).

Results from theoretical biology
~\citep{McNamara2008, Campenni2014, Debove2015b, Ecoffet2020}
have shown that cooperation with partner choice is optimal only under certain conditions. First, the number of cooperation opportunities must be large enough that an agent can refuse to cooperate with a potential partner and still have the opportunity to meet a more interesting partner. Second, if an agent and its partner both decide to cooperate, the actual duration of this cooperation must be long enough to make cooperation with an uninteresting partner significantly costly (which is the case when there can be only one single successful cooperation event). Under these conditions, the optimal strategy for an agent is to be very demanding in choosing its partner.%, which can only be possible if the probability of meeting this "ideal" partner is non-zero.

The question raised in this paper is whether reinforcement learning algorithms actually succeed in learning an optimal strategy when the necessary conditions are met. We are particularly interested in how the rarity of significant events influences convergence speed and performance of policy learning. Indeed, it is not clear how gradient-based policy search method can deal with a possibly large number of non-significant events that provide zero-reward.

We use two state-of-the-art methods for on-policy reinforcement learning with continuous state and action spaces: (1) a deep learning method (PPO~\citep{Schulman2017}) for gradient policy search and (2) an evolutionary method (CMAES~\citep{Hansen2001}) for direct policy search. While both methods provide similar results when the agent is always presented with significant events, policy search methods are not equals when such events become rarer. While the direct policy method is oblivious to rarity of significant events, the gradient policy search method suffers significantly from rarity. 

The paper is structured as follows: the reinforcement learning problem with significant rare events and single reward per episode is formalized, and the partner choice learning problem is presented as a variation of a continuous prisoner's dilemma. Algorithms and results are then presented, and learned policies are analysed and compared.

\section{Methods}

\subsection{Learning with Rare Significant Events}\label{sec:rarerewards}

Formally, we consider an independent learner $x_{\bullet}$, called the \textit{focal agent}, which is placed in an aspatial environment. At each time step, $x_{\bullet}$ is presented with either a \textit{cooperative partner} $x_i^+ \in X^+$ or a \textit{non-cooperative partner} 
$x_j^- \in X^-$. $X^+$ (resp. $X^-$) is the finite set of all cooperative (resp. non-cooperative) agents, with both $i$ and $j \in \mathbb{N}$ and $i>0,j\geq0$. 
When presented with a non-cooperative partner $x_j^-$, the focal agent's reward will always be zero. When presented with a cooperative partner $x_i^+$, the focal agent's reward will depend on its own action and that of its partner. 
%When presented with a cooperative partner $x_i^+$, the focal agent's reward can be different from zero
%positive or null => en fait peut aussi être négatif
%non-zero
(see Section~\ref{ssec:payoff} for details). 

Our objective is to endow the focal agent $x_{\bullet}$ with the ability to learn how to best cooperate, which implies to negotiate with its potential partners and decide whether cooperation is worth investing energy in, or not (see Section~\ref{ssec:negociation} for details). The focal agent faces an individual learning problem as it must optimize its own gain over time in a competitive setup,  whether its partners are also learning agents or not. 
For cooperation to occur between the focal agent and a partner, the partner must willing to cooperate (ie. be one of $x_i^+$) and both the focal agent \textit{and} the cooperative partner must estimate that one's own energy invested in cooperation is worth the benefits.

We use the standard reinforcement learning framework proposed by~\citet{sutton2018reinforcement} to formalize the learning task from the focal agent's viewpoint, which is essentially a single agent reinforcement learning problem. 

The focal agent $x_{\bullet}$ interacts with the environment in a discrete time manner. At each time step $t = 0, 1, 2, ...$, $x_{\bullet}$ is in a state $s \in \mathbb{R}$ which describes its current partner's investment value, and plays a continuous value $a \in \mathbb{R}$ which represents its decision to cooperate ($a>0$) or not ($a<=0$).

Let $\pi_\theta$ be the parametrised policy of the focal agent, with $\theta \in \mathbb{R}^n$. The learning task is to search for $\theta^*$, such as:

\begin{equation}
    \theta^* = \underset{\theta}{argmax} J(\theta)
    \label{eq:learning}
\end{equation}

With $J$ the global function to be optimized, defined as:

\begin{equation}
    J(\theta) = \mathbb{E}\underset{t}{\sum} r_t
    \label{eq:J}
\end{equation}

with reward $r_t$ 
%(or return)
at time $t$. Rewards are defined such that $r \in \mathbb{R}$ and depends on the current state $s$ and action $a$, and are produced according to the probability generator defined as follow:

\begin{equation}
    r(s,a) = \left\{
        \begin{array}{ll}
            \mathit{payoff}(s,a) & \mbox{with probability } p \\
            0 & \mbox{otherwise.}
        \end{array}
    \right.
    \label{eq:rewards}
\end{equation}

The probability $p\in [0,1]$ determines the probability to encounter a cooperative agent (i.e. one of $x_i^+$). The value of $p$ depends on the setup, and determines how \textit{rare} significant events occur when $p<1.0$. A probability of $p=1.0$ means the focal agent $x_{\bullet}$ encounters a cooperative partner at each time step $t$, with a possible positive reward (if cooperation is accepted by both agents) that depends on the $\mathit{payoff}$ function. Non-zero rewards become rarer (but still possible) as $p \rightarrow 0$. Note that $\mathit{payoff}(s,a)$ is non-zero \textit{only} if both the focal agent \textit{and} its cooperative partner accept to cooperate. Cf. Section~\ref{ssec:negociation} for details on the negotiation process.

The problem presented here is very similar to that of Rare Significant Events as formulated by~\citet{frank2008}. However, our problem differs on two aspects. Firstly, we consider on-line on-policy search of a parametrised policy, where the frequency of significant events cannot be controlled. Secondly, and even more importantly, a learning episode stops right after the focal agent and one cooperative agent have reached a consensus to cooperate. If no cooperation is triggered, an episode stops after a maximum number of iterations $T$, defined as:

\begin{equation}
    T = \frac{100}{p} \mbox{ time steps}
    \label{eq:T}
\end{equation}

It results that the expected number of meetings $M$ is held constant independently from the value of $p$ (i.e. $\mathbb{E}(M) = 100$). It is therefore possible to obtain episodes of different lengths but with the same number of significant events.

The situation that is modelled here corresponds to many collective tasks observed in nature \citep{Bshary2003, Simms2002, Wilkinson2016}, where each agent has to balance between looking for partners and cooperating with the current partner, the latter possibly taking significant time. As a matter of fact, it has been shown elsewhere
~\citep{McNamara2008, Campenni2014,Debove2015b,Ecoffet2020,Ecoffet2020b} 
that optimal partner choice strategies can be reached only when the cost of cooperation is large (ie. the duration of cooperation is long with regards to looking for cooperative partners).

% =-=-=-=-= =-=-=-=-= =-=-=-=-= =-=-=-=-= =-=-=-=-= =-=-=-=-= 
% =-=-=-=-= =-=-=-=-= =-=-=-=-= =-=-=-=-= =-=-=-=-= =-=-=-=-= 
% =-=-=-=-= =-=-=-=-= =-=-=-=-= =-=-=-=-= =-=-=-=-= =-=-=-=-= 

\subsection{Partner Choice and Payoff Function} \label{ssec:payoff}

Whenever the focal agent $x_\bullet$ and a cooperative partner $x_i^+$ interact together, they play a variation of a continuous Prisoner's Dilemma. Cooperation actually takes place if \textit{both} agents deem it worthwhile. The two-step procedure for partner choice is the following:

\begin{enumerate}
    \item each agent simultaneously announce the \textit{investment} they are willing to pay to cooperate;
    \item each agent then \textit{chooses} to continue the cooperation based on the investment announced by its partner and its own. 
\end{enumerate}

% The focal agent invests the value $cost(x_\bullet)$ and the partner $cost(x_i^+)$.
To simplify notations, we use $x_\bullet$ and $x_i^+$ to represent both the agents and the investment values they play, i.e. $x_\bullet$ (resp. $x_i^+$) plays $x_\bullet$ (resp. $x_i^+$). The gain received by the focal agent $x_\bullet$ is defined as:

\begin{equation}
    P(x_\bullet, x_i^+) = a \times x_\bullet + b \times x_i^+ - \frac{1}{2} x_\bullet^2
    \label{eq:extendedPD}
\end{equation}

With $a,b \geq 0$ and $a+b>0$. This payoff function combines both a prisoner's dilemma and a public good game, and was first introduced in~\citet{Ecoffet2020}. 
Two different equilibria\footnote{These are actually Nash equilibria, when all agents are learning.} can be reached for $x_\bullet$:

\begin{itemize}
    \item $x_d = a$. This is a sub-optimal equilibrium, which corresponds to an agent cheating, a typical outcome in the prisoner's dilemma where an agent maximizes its own gain, but also minimizes its exposure to defection. This ensure the best payoff for the agent if it is unable to distinguish a cheater from a cooperator.
    \item $x_c = a + b$. This is the optimal equilibrium, where both agents cooperate to maximize their long-term gain.
\end{itemize}

The public good game is included in the payoff function to help distinguish between agents that are simply ignoring the cooperation game ($x_\bullet=0$), from those who takes part in it, even if they defect  ($x_\bullet \geq x_d$).

The focal agent can get the optimal payoff if it plays $x_\bullet = x_c$ \textit{and} its partner plays $x_i^+ \geq x_c$, which can occur if particular conditions are met when partner choice is enabled. Partner choice can lead to optimal individual gain whenever a successful cooperation removes the possibility for further gain with other partners. In other words: the focal agent can meet with any number of possible partners but will take the gain of the first and single mutually accepted cooperation offer.

In this paper, we set $a = 5$ and $b = 5$, therefore $x_d = 5$ and $x_c = 10$. The maximum payoff the agent can obtain is to invest $x_\bullet = x_c$ with its partner investing equally $x_i^+ = x_c$. In this context, $P(x_\bullet, x_i^+) = 50$. The focal agent's investment is bounded as $0.0 \leq x_\bullet \leq 15.0$. This is similar for $x_i^+$.

$P(x_\bullet, x_i^+)$ and $\mathit{payoff}(s,a)$ (introduced in Equation~\ref{eq:rewards}) differs as the $P$ function relates to the game theoretical setting while the $\mathit{payoff}$ function relates to the reinforcement learning problem. On the one hand, the $\mathit{payoff}$ function computes the focal individual's reward whether \textit{or not} cooperation was initiated. On the other hand, $P$ computes the focal individual's gain that results from a cooperation game between two agents that \textit{accepted} to cooperate. However, both functions are linked. From a notational standpoint, $s$ represents the investment value of the focal individual $x_\bullet$, and $a$ represents the decision to cooperate and depends on both $s$ and that of its partner $s_i^+$ (which is implicit). The return value of $\mathit{payoff}(s,a)$ depends on whether cooperation was initiated or not. If both agents decided to cooperate, then the focal agent's payoff is $\mathit{payoff}(s,a) = P(x_\bullet, x_i^+)$, with $P(x_\bullet, x_i^+) \leq 50$ in this case. If cooperation fails, the focal agent's payoff is $\mathit{payoff}(s,a) = 0$ (which is obtained without having to compute $P$). The $payoff$ function in Equation~\ref{eq:rewards} can be written as follow, with updated notations and assuming $a_\bullet>0$ (resp. $a_i^+>0$) means the focal agent (resp. partner) is willing to cooperate:

\begin{equation}
    \mathit{payoff}(s_\bullet,a_\bullet) = \left\{
        \begin{array}{ll}
            P(x_\bullet,x_i^+) & \mbox{if $a_\bullet>0$ and $a_i^+>0$} \\
            %P(x_\bullet,x_i^+) & \mbox{if both} a_\bullet \mbox{and} a_i^+ \mbox{are positive}\\
            0 & \mbox{otherwise.}
        \end{array}
    \right.
    \label{eq:rewards_expanded}
\end{equation}

% =-=-=-=-= =-=-=-=-= =-=-=-=-= =-=-=-=-= =-=-=-=-= =-=-=-=-= 
% =-=-=-=-= =-=-=-=-= =-=-=-=-= =-=-=-=-= =-=-=-=-= =-=-=-=-= 
% =-=-=-=-= =-=-=-=-= =-=-=-=-= =-=-=-=-= =-=-=-=-= =-=-=-=-= 

%\subsection{Decision-Making}

%The agent makes two decisions in sequence at each time step. The agent first picks its investment value, then it gets as an observation its partner investment and can decide whether it will interact with it. These two decisions are split into two different decision modules: The investment module and the partner choice module. The investment module determines how much the agent is willing to invest, it takes no observation, 

\subsection{Behavioural Strategies}\label{ssec:negociation}

For each interaction, the focal agent's investment value $x_\bullet \in [0,15]$ is computed, and when the investment value of its partner is known, its decision to cooperate $a_\bullet \in \mathbb{R}$ is computed to determine if cooperation should be pursued or not. Each value is provided by a dedicated decision module:

\begin{itemize}
    \item the \textbf{investment module} which provides the cost $x_\bullet$ that the focal agent is willing to invest to cooperate. This module takes no input as it is endogenous to the agent (i.e. the proposed cost $x_\bullet$ is fixed throughout an episode);
    \item the \textbf{choice module} takes both the focal agent's own investment value ($x_\bullet$) and that of its partner ($x_i^+$ or $x_j^-$), and computes $a_\bullet$, which is used to determine if cooperation is an interesting choice ($a_\bullet > 0$) or not ($a_\bullet \leq 0$). The choice module is essentially a function $f_{choice}(x_\bullet,x_{partner}) \rightarrow a_\bullet$ with $x_{partner} \in X^+ \cup X^-$. The parameters of the function are learned, and the decision to cooperate is computed (as the decision to cooperate is conditioned by the partner's investment).
\end{itemize}

With respect to the focal individual, Section~\ref{sec:algo} describes how the investment and choice modules are defined and how learning is performed depending on the learning algorithm used. 

Cooperative partners $x_i^+$ and non-cooperative partners $x_j^-$ also use similar decision modules, providing investment and choice values. However, all use deterministic fixed strategies, which may differ from one partner to another. Firstly, non-cooperative partners $x_j^-$ all follow the same strategy. Both the investment value $x_j^-$ and the decision to cooperate $a_j^-$ are always $0$, $\forall j$. 

Secondly, cooperative partners $x_i^+$ each follows a stereotypical cooperative strategy depending on the value $i$. Each cooperating partner invests a fixed value $x_i^+ \in [0, 15]$ defined as:

\begin{equation}
    x_i^+ = \frac{i-1}{i_{max}} \times 15,\,i \in \{1, \ldots, i_{max}\} 
    \label{eq:investx+}
\end{equation}

Each cooperative partner then accepts to cooperate if the focal agent's investment value $x_\bullet$ is greater or equal to their investment, which is written as follow:

\begin{equation}
        a_i^+ =\left\{
        \begin{array}{ll}
            1 & \mbox{if $x_\bullet \geq x_i^+$} \\
            %P(x_\bullet,x_i^+) & \mbox{if both} a_\bullet \mbox{and} a_i^+ \mbox{are positive}\\
            -1 & \mbox{otherwise.}
        \end{array}
    \right.
    \label{eq:acceptx+}
\end{equation}

In the following, there are $i_{max}=31$ cooperating partners ($x_i^+ \in X^+, i \in \{1, \ldots, 31\}$). Following Eq.\ref{eq:acceptx+}, this means cooperating partner $x_1^+$ (resp. $x_2^+$, ..., $x_{31}^+$) plays $0$ (resp. $0.5$, ..., $15$).

% =-=-=-=-= =-=-=-=-= =-=-=-=-= =-=-=-=-= =-=-=-=-= =-=-=-=-= 
% =-=-=-=-= =-=-=-=-= =-=-=-=-= =-=-=-=-= =-=-=-=-= =-=-=-=-= 
% =-=-=-=-= =-=-=-=-= =-=-=-=-= =-=-=-=-= =-=-=-=-= =-=-=-=-= 

% =-=-=-=-= =-=-=-=-= =-=-=-=-= =-=-=-=-= =-=-=-=-= =-=-=-=-= 
% =-=-=-=-= =-=-=-=-= =-=-=-=-= =-=-=-=-= =-=-=-=-= =-=-=-=-= 
% =-=-=-=-= =-=-=-=-= =-=-=-=-= =-=-=-=-= =-=-=-=-= =-=-=-=-= 

\section{Parameter Settings and Algorithms} \label{sec:algo}

We use two reinforcement learning algorithms: a gradient policy search algorithm (PPO) and a direct policy search algorithm (CMAES). Both algorithms are used to learn the parameters of the focal agent's decision modules.

For both algorithms, the performance of a policy (i.e. the \textit{return} or the \textit{fitness}, depending on the vocabulary used) during one episode is computed as the sum of rewards during the episode (cf. Section~\ref{sec:rarerewards}), which is either zero, or the value of the unique non-zero reward obtained before the episode ends. 

%\subsection{Generalities}

%Je colle cette section là, mais il faut la faire disparaître: à discuter en visio. Grandes lignes:
%- figures a mettre en annexe, ou a condenser dans une seule figure (OK on changeant les textes par les notations).
%- répartir les détails dans techniques dans les sections PPO et CMAES
%- ajouter un tableau pour CMAES résumant les paramètres (nombre d'éval, etc.) y compris nombre de poids.
%- manque aussi le nombre de dimension (poids) du problème dans le tableau de PPO 
%- une fois tout cela résolu on verra si on garde une section dédiée pour les algos (pour l'instant oui)
%- remarque sur le nombre de dimension: PPO en a plus. Ca peut expliquer la lenteur de convergence (je pense pas, mais c'est une critique qu'on peut faire). Moyen de corriger cela: augmenter artificiellement les dimensions du problème pour CMAES (soit en mettant des dimensions dummy, soit en disons qu'un paramètre c'est deux parties du génome qu'on additionne). A mettre en annexe

% =-=-=-=-= =-=-=-=-= =-=-=-=-= =-=-=-=-= =-=-=-=-= =-=-=-=-= 
% =-=-=-=-= =-=-=-=-= =-=-=-=-= =-=-=-=-= =-=-=-=-= =-=-=-=-= 
% =-=-=-=-= =-=-=-=-= =-=-=-=-= =-=-=-=-= =-=-=-=-= =-=-=-=-= 

\subsection{Proximal Policy Optimization}

The deep reinforcement learning Proximal Policy Optimisation (PPO) \citep{Schulman2017} is a variation of the Policy Gradient algorithm \citep{sutton2018reinforcement}. Policy gradient algorithms maximize the global performance by updating the parameters $\theta$ of the policy $\pi$ (cf. Eq.~\ref{eq:J}).

Though, as the expected value of a certain state-action pair varies according to the policy itself, updating a new policy from samples acquired from an old policy may cause inaccurate predictions, as the expected value of an action-state pair may be wrong with respect to the new policy. PPO ensures that the policy generated from the samples of the new policy does remain in a so-called trust region at each learning step.
%This ensures the stability of the learning process. % Commentaire d'OS: c'est fait grace au terme de KL divergence.

As we are dealing with episodes and do not want to encourage the focal agent to act in the least amount of time steps as possible, the discount factor is set to $\gamma = 1.0$, as recommended by~\citet[p.68]{sutton2018reinforcement}. The PPO hyper-parameters used are reported in Table~\ref{tab:ppo_parameters}.

\begin{table}[tbp]
    \centering
    \begin{tabular}{cc}
        \toprule
        \textbf{Parameters} & \textbf{Values} \\
        \midrule
        Learning rate & $0.005$\\
        Optimiser Algorithm & SGD \\
        Number of optimisation epochs & $10$\\
        Minibatch size & $128$ \\
        Batch size & $4000$ \\
        %Kullback-Leibler coefficient & 0.2 \\
        %Kullback-Leibler target & 0.01 \\ % => commentaire OS: dans la Table, si on met KL, il faudrait donner l'expression de la fonction optimisée avec le terme KL pour qu'on comprenne les params.
        %GAE Parameter $\lambda$ & 1.0 \\
        Discount factor $\gamma$ & 1.0 \\
        %Value Function loss coefficient & 1.0 \\
        %Entropy coefficient & 0.0 \\ % should be: 0 to 0.01
        Search space PPO-MLP ($\theta_{MLP}$) & $\mathbb{R}^{33}$\\
        Search space PPO-DEEP ($\theta_{DEEP}$) & $\mathbb{R}^{133894}$\\
        \bottomrule
    \end{tabular}
    \caption{Parameters for the PPO algorithm}
    \label{tab:ppo_parameters}
\end{table}

The investment and choice modules are both represented as Artificial Neural Networks (ANN). A module is composed of both a decision network and a Value function, as PPO runs as an actor-critic algorithm. The Value function network has the same layout as the decision network, but only output the (continuous) value of the state. 

The decision network for the investment module is a simple neural network 
% INTERNAL COMMENT ABOUT THE INVESTMENT MODULE ARCHITECTURE 1/2
% 2021-01-13: for clarity, I consider as implementation detail the sentence below, as one dummy input to 0.0 means associated weights are *meaningless*.
%=> "with one dummy input (always 0.0, due to implementation constraints of the library used) and a bias (always 1.0),"
% See CMAES section for further comments on this.
with one single input set to $1.0$, no hidden layer and two outputs: the investment mean $m$ and standard deviation $\sigma$. The investment $x_\bullet$ is picked along the distribution $\mathcal{N}(m, \sigma^2)$ and clipped between 0 and 15. The continuous stochastic action selection is essential to the PPO search algorithm. %The neural network has $4$ weights, and the value function has $2$ weights. In total, the investment module is composed of $6$ weights.

The decision network for the choice module is a multilayer perceptron with two input neurons and two output neurons (for accepting or refusing cooperation). The output neurons use a linear activation function, and a softmax probabilistic choice is done to choose which action to make (accept or decline). Hidden units use an hyperbolic tangent activation function. A bias node is used, that projects on both the hidden layer(s) and output neurons. The Value Function estimator use the same architecture as the choice neural networks, with only one output.

In Section~\ref{sec:results}, two different architectures are evaluated, which we refer to as \textbf{PPO-MLP} and \textbf{PPO-DEEP}. While both use the decision network for the investment module described before, they differ with respect to the architecture used for the choice module. PPO-MLP implements a single hidden layer with 3 neurons, and PPO-DEEP implements a deep architecture with two hidden layers, each with $256$ neurons. While PPO-DEEP may seem overpowered at first sight, over-parametrization has been shown to be very effective in deep learning as multiple gradients can be followed in wide neural networks~\cite{du2018icml,neyshabur2018arxiv,allen2019convergence}.

% 
% * PPO-MLP et PPO-DEEP Investment module
% ** 2 paramètres (decision) + 1 parametre (value fn)
% * choice module:
% ** PPO-MLP: 
% *** decision: ( 2 entrées + 1 biais ) * 3 + ( 3 + 1 biais ) * 2 sorties, soit 17 paramètres
% *** value function: ( 2 entrées + 1 biais ) * 3 + ( 3 + 1 biais ) * 1 sortie, soit 13 paramètres
% ** PPO-DEEP: 
% *** decision: ( 2 entrées + 1 biais ) * 256 + ( 256 + 1 biais ) * 256 + ( 256 + 1 biais ) * 2 sorties, soit 67074 paramètres
% *** value function: ( 2 entrées + 1 biais ) * 256 + ( 256 + 1 biais ) * 256 + ( 256 + 1 biais ) * 1 sortie, soit 66817 paramètres
% *
% * => PPO-MLP: 33 paramètres
% * => PPO-DEEP: 133894 paramètres

All parameter values and module architecture result from an extensive search (summarised in the Supplementary Materials).
% ANONYMOUS
%, Section~\ref{supmat}). 
In particular, a grid search was performed to select the best values for each parameters, including the learning rate ($lr$). The number of Simple Gradient Descent iterations, the batch size and the mini-batch size had little impact on neither performance nor convergence. In addition, we performed additional experiments to evaluate the impact of using (1) a discount factor $\gamma<1.0$ (i.e. $0.9$, $0.99$ and $0.999$) and (2) PPO without actor-critic. None of these settings provided better (or even comparable) results to those obtained with the parameters used in Table~\ref{tab:ppo_parameters}.

% =-=-=-=-= =-=-=-=-= =-=-=-=-= =-=-=-=-= =-=-=-=-= =-=-=-=-= 
% =-=-=-=-= =-=-=-=-= =-=-=-=-= =-=-=-=-= =-=-=-=-= =-=-=-=-= 
% =-=-=-=-= =-=-=-=-= =-=-=-=-= =-=-=-=-= =-=-=-=-= =-=-=-=-= 

\subsection{Covariance Matrix Adaptation Evolution Strategy}

The Covariance Matrix Adaptation Evolution Strategy (CMAES) is an optimisation algorithm that does black box optimisation and is derivative-free \citep{Hansen2001}. The goal of CMAES is to find $\theta^*$ that maximizes (or minimizes) a continuous function $f$. CMAES does not require the function to be convex or differentiable, and relies on stochastic sampling around the current estimate of the solution. CMAES creates a population of size $\lambda$ using a multivariate Gaussian distribution. Each individual of the population is evaluated and CMAES then updates its distribution estimation based on the average of the sampled agents weighted by their evaluation rank. Furthermore, the covariance matrix of the multivariate Gaussian distribution is updated so that the distribution is biased toward the most promising direction.

The investment module is represented as a single real value (the investment), which is clipped between 0 and 15 when used. 
% INTERNAL COMMENT ABOUT THE INVESTMENT MODULE ARCHITECTURE 2/2 
% Following sentence removed as originally used for implementation details and turned out to being useless : "An additional "dummy" real value is added, which is optimized but not used, to enforce a similar number of dimensions with that of the PPO investment module." 
% Additional comment: for both PPO and CMAES, the extra 1 to 3 parameters which are removed from the text here were actually not used as not connected to any function (ie. any values is possible, investment module will not be affected). We also consider *negligible* these 1-3 extra dimensions w.r.t. convergence speed.
The partner choice module is a neural network with 2 inputs, one hidden layer with three neurons and two neurons on the output layer used to compute the probability to \textit{accept} or \textit{refuse} cooperation. A softmax probabilistic choice is made to choose which action to make. A bias node is also used, neurons from the hidden layer use an hyperbolic tangent activation function, and the output units use a linear activation function. There are $17$ neural network weights. 

The parameters for both modules are compiled into a single vector of real values. To make the search space similar to that of PPO, dummy parameters are added to the vector (i.e. values which can be modified by the algorithm, but with no impact on the outcome) to reach a total number of $34$ real values (i.e. $\Theta\in\mathbb{R}^{34}$).

Table~\ref{tab:cmaparam} summarizes the parameters used for the CMAES algorithm. As CMAES is mostly parameter-free, there were no need to perform extensive preliminary search, and we used the default values. We choose $\sigma_{init} = 1.0$ for the initial standard deviation and a vector of zeros as initial guess. The population size $\lambda$ is the default population size in the python CMAES implementation \citep{pycma}, i.e. $\lambda = 4+\floor{3\times\ln(N)} = 14$ with $N$ the number of dimensions in the model.
%We encode the weights of both modules into one genome composed of 34 genes (a "candidate solution" of the focal agent). 
Once the $\lambda$ candidate solutions are evaluated, a new population is generated according to their performance. A new population is generated every 14 episodes, and so forth until the evaluation budget is consumed. 

A candidate solution for the focal agent is evaluated on one episode only, which length may vary depending on when the focal agent and its partner both accepts to cooperate (maximal duration defined in Eq.~\ref{eq:T}).

\begin{table}
    \centering
    \begin{tabular}{cc}
    \toprule
        \textbf{Parameter} &  \textbf{Value} \\
     \midrule
        Population size & 14 \\
        Number of episode per evaluation & 1 \\
        $\sigma_{init}$ & 1.0 \\
        Search space ($\theta_{CMAES}$) & $\mathbb{R}^{34}$ \\
     \bottomrule
    \end{tabular}
    \caption{Parameters for the CMAES algorithm}
    \label{tab:cmaparam}
\end{table}

% =-=-=-=-= =-=-=-=-= =-=-=-=-= =-=-=-=-= =-=-=-=-= =-=-=-=-= 
% =-=-=-=-= =-=-=-=-= =-=-=-=-= =-=-=-=-= =-=-=-=-= =-=-=-=-= 
% =-=-=-=-= =-=-=-=-= =-=-=-=-= =-=-=-=-= =-=-=-=-= =-=-=-=-= 

\section{Results}
\label{sec:results}

The environment, the models and the learning algorithms are implemented with ray\footnote{\url{https://docs.ray.io/en/master/}}, rllib\footnote{\url{https://docs.ray.io/en/master/rllib.html}} and pytorch\footnote{\url{https://pytorch.org/}}. We use the cma\footnote{\url{https://pypi.org/project/cma/}} package in python for the CMAES implementation. Source code is available at \url{https://github.com/PaulEcoffet/RLCoopExp/releases/tag/v1.1}.

For a given value of probability of rare significant events $p$, we performed $24$ independent runs for each algorithm. A run lasts $200\,000$ episodes. The maximum duration of an episode is fixed as described in Section~\ref{sec:rarerewards} so the expected number of significant events remains identical independently from the \textit{actual} rarity throughout one episode (cf. equation~\ref{eq:T}). In practical, an episode lasts \textit{at most} $100$ (resp. $200$, $500$, $1000$) iterations for $p=1.0$ (resp. $0.5$, $0.2$, $0.1$). 

Performance of the current policy is plotted every $4000$ iterations, which corresponds to the batch size used by both PPO instances for learning. As episodes last significantly shorter than $4000$ iterations this means the policy's performance is averaged. 
For CMAES, we extract the best policy of the current generation and re-evaluate it $10$ times (i.e. for $10$ episodes) to get a similarly averaged performance. Results are shown on figures with a data point every $1000$ episodes.

% precision:
% PPO: a plotted data point is a compilation of the average performance of each batch (ie. 4000 iterations, variable number of episodes) for as many batch it is required to reach 1000 episodes, taking an episode every 10 episodes (ie. actual number of batches may vary)
% CMAES: a plotted data point is a compilation of the best policies taken every 100 episodes for the last 1000 episodes (ie. 100 values), with one individual's performance resulting from 10 reevaluations (reduce variance).

% =-=-=-=-= =-=-=-=-= =-=-=-=-= =-=-=-=-= =-=-=-=-= =-=-=-=-= 
% =-=-=-=-= =-=-=-=-= =-=-=-=-= =-=-=-=-= =-=-=-=-= =-=-=-=-= 
% =-=-=-=-= =-=-=-=-= =-=-=-=-= =-=-=-=-= =-=-=-=-= =-=-=-=-= 

\subsection{Learning when All Events are Significant}

\begin{figure}
    \centering
    \includegraphics{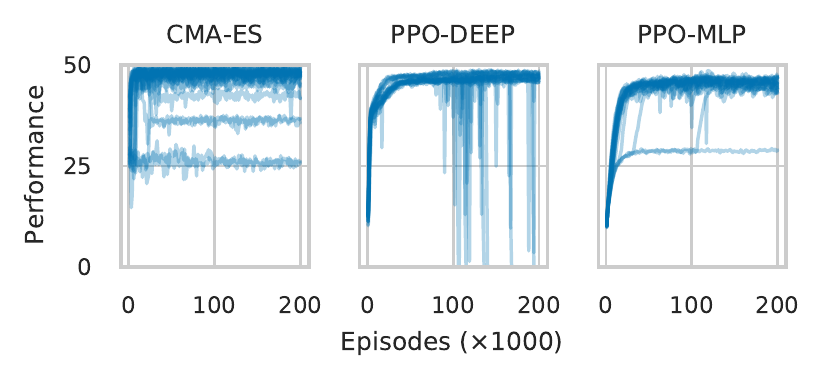}
    \caption{Performance of the best policy throughout learning with CMAES (top), PPO-DEEP (center) and PPO-MLP (bottom), with 24 independent runs per method, for $200*10^3$ episodes. There are $20/24$ runs that produced a policy where performance above $40$ with CMAES, $20/24$ for PPO-DEEP and $23/24$ for PPO-MLP. Note that PPO-DEEP produces $24/24$ runs with performance above $40$ around episode $80*10^3$, with performance occasionally degrading and immediately recovering for some runs afterwards due to the learning step size (see Annex for further analysis).}
    \label{fig:all_algo_reward_mean_detail_p1}
\end{figure}

Figure~\ref{fig:all_algo_reward_mean_detail_p1} shows the performance throughout learning for CMAES, PPO-DEEP and PPO-MLP when $p=1.0$ (i.e. the focal agent faces only cooperative partners). Each Figure shows $24$ curves corresponding the $24$ independent runs. Both PPO versions and CMAES are shown to learn near optimal policies ($performance \rightarrow 50$) in almost all runs. CMAES is the fastest to converge, and PPO-DEEP (despite the huge number of dimensions) is faster than PPO-MLP. On the other hand, CMAES offers less robustness as $20$ (out of $24$) runs with CMAES reach a performance above $40$, to be compared to $23$ (out of $24$) runs with PPO-MLP and $24$ runs with PPO-DEEP.
%The case of PPO-DEEP is particular: all $24$ runs converge to a performance above $40$ before episode $75\bullet10^3$, and $TODO$ runs then show degraded performance, possibly due to misleading gradients (see also Annex).

\begin{figure}
    \centering
    \includegraphics{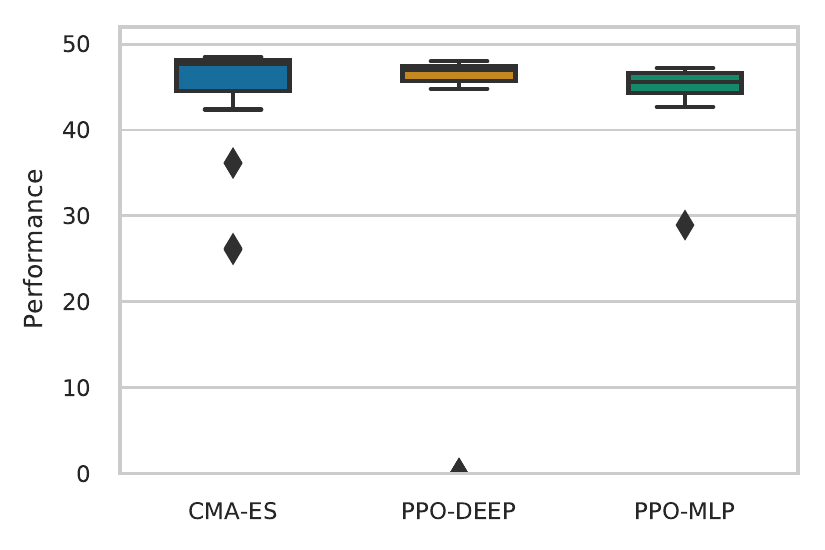}
    \caption{Performance of the best policies from CMAES, PPO-DEEP and PPO-MLP with $p = 1.0$  after re-evaluating policies for $1000$ episodes without learning. Two-tailed Mann-Whitney U-test, $n=24$, gives $p-value = 0.12$ (CMAES vs. PPO-DEEP), $p-value = 0.019$ (CMAES vs. PPO-MLP), $p-value = 0.018$ (PPO-DEEP vs. PPO-MLP). Median values and Median Absolute differences are:
    CMAES (median=$47.64$, MAD=$4.72$) is only slightly more than that of PPO-DEEP (median=$46.99$, MAD=$13.04$) and PPO-MLP (median=$45.58$, MAD=$1.86$).} % TODO: U test a mettre à jour pour les *3*
    \label{fig:postmortemcompreward}
\end{figure}

In order to better compare the quality of the policies learned by each algorithm, the best policy from the end of each run is selected and re-evaluated for $1000$ extra episodes without learning. Results are shown in Figure~\ref{fig:postmortemcompreward} with all three methods faring similar performance. The median value for CMAES ($47.64$) is only slightly more than that of PPO-DEEP ($46.99$) and PPO-MLP ($45.58$).

%The mean value for CMAES ($44.42 \pm 6.62$) is only slightly less than that of PPO-MLP ($44.79 \pm 3.57$), PPO-DEEP ($39.12 \pm 17.89$)
%and while the difference between the two distribution may be interpreted as significant ($\text{\emph{p}}$-values > $0.01$, Mann-Whitney's U-test).%, the effect size is peculiarly low ($d = 0.07$, Cohen's term). 
Therefore, we conclude that all three algorithms provide excellent and comparable results when only significant events are experienced ($p=1.0$).

% =-=-=-=-= =-=-=-=-= =-=-=-=-= =-=-=-=-= =-=-=-=-= =-=-=-=-= 
% =-=-=-=-= =-=-=-=-= =-=-=-=-= =-=-=-=-= =-=-=-=-= =-=-=-=-= 
% =-=-=-=-= =-=-=-=-= =-=-=-=-= =-=-=-=-= =-=-=-=-= =-=-=-=-= 

\subsection{Learning when Significant Events are Rare}

%We investigate the impact of the rarity of rewards on the agent's performance. The expected number of meeting of $x^+$ partners for a whole episode is held constant, $E(M) = 100$, for all $p$, thus the expected reward should not differ between conditions due to a more restrained choice depending on $p$.

\begin{figure}
    \centering
    \includegraphics{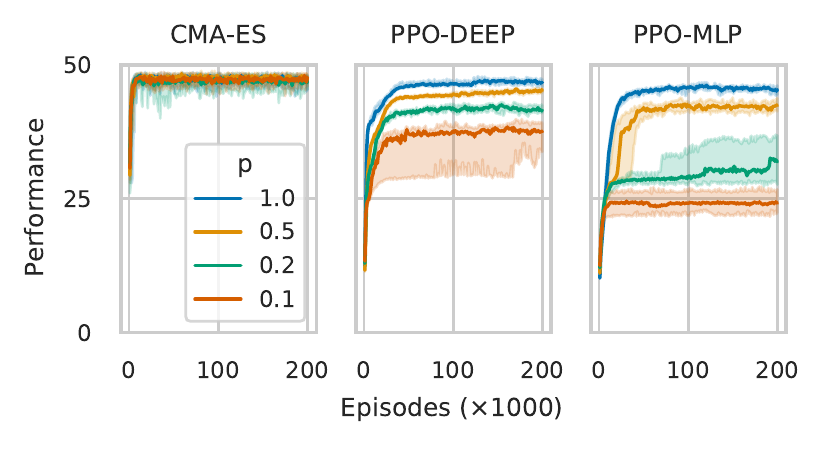}
    \caption{Performance of the best policies (median and $95 \%$ confidence interval) throughout learning with CMAES, PPO-DEEP and PPO-MLP for the 3 conditions with rare significant events $(p \in \{0.1,0.2,0.5\})$ and 1 control condition ($p=1.0$, same data as shown in Fig.\ref{fig:all_algo_reward_mean_detail_p1}), for the first $75*10^3$ episodes (out of $200*10^3$).}
    \label{fig:all_algo_reward_all_p}
\end{figure}

Figure~\ref{fig:all_algo_reward_all_p} show the performance of the agent throughout its learning with both PPO algorithms and the CMAES algorithm for different conditions of rare significant events ($p \in \{0.1, 0.2, 0.5\}$), as well as with the control condition when all events are significant ($p=1.0$, taken from the previous Section). Each figure shows the mean performance of $24$ independent runs per conditions, compiling each setup by tracing the median performance and 95\% confidence interval from the $24$ runs. 

CMAES is only marginally impacted when significant events become rarer (i.e. $p<1.0$), with all setups showing convergence towards a similar performance value close to the optimal (above $40$). While PPO-DEEP fares better than PPO-MLP for $p<1.0$, both are largely affected. In the extreme case where $p = 0.1$, the average performance of $35.7\pm5.2$ for PPO-DEEP and $24.9\pm4.2$ of PPO-MLP, to be compared to $46.2\pm3.2$ for CMAES.

\begin{figure}
    \centering
    \includegraphics{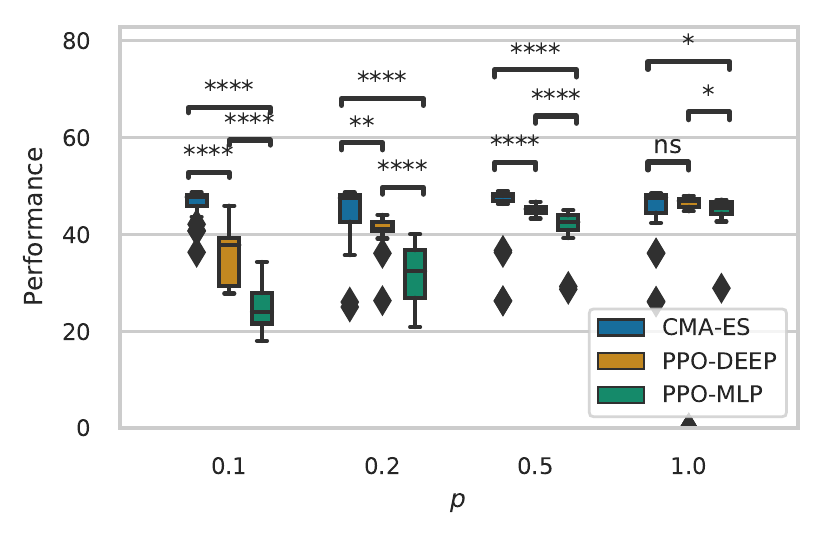}
    \caption{Performance of the best policies (medians and quartiles) from CMAES, PPO-DEEP and PPO-MLP with $p \in \{0.1, 0.2, 0.5, 1.0\}$ after re-evaluating policies for $1000$ episodes without learning. Two-tailed Mann-Whitney U-test, $n=24$ marked as: * for $p-value < 0.05$, ** for $p-value < 0.01$, *** for $p-value < 0.001$ and **** for $p-value < 0.0001$.}
    \label{fig:postmortemcomprewardfull}
\end{figure}

Figure~\ref{fig:postmortemcomprewardfull} shows the results for the additional analysis where the best policy from each run for each condition $p \in \{0.1, 0.2, 0.5, 1.0\}$ is selected and re-evaluated for $1000$ extra episodes without learning and with the condition $p=1.0$ (i.e. only significant events matter here). Results confirm that the difference in the performance of policies obtained with CMAES compared to either versions of PPO widens as significant events become rarer ($p < 1.0$) with both PPO-MLP and PPO-DEEP faring significantly worse than CMAES ($\text{\emph{p}}$-value $ < 0.0001$, Mann-Whitney U-test). % TODO même p-value pour les deux j'imagine? Oui sauf pour p=0.2 entre PPO-DEEP et CMA-ES (p < 0.01)

% =-=-=-=-= =-=-=-=-= =-=-=-=-= =-=-=-=-= =-=-=-=-= =-=-=-=-= 
% =-=-=-=-= =-=-=-=-= =-=-=-=-= =-=-=-=-= =-=-=-=-= =-=-=-=-= 
% =-=-=-=-= =-=-=-=-= =-=-=-=-= =-=-=-=-= =-=-=-=-= =-=-=-=-= 

\subsection{Analysing the Best Policies for Partner Choice}

In order to better understand why policies' performance differ among learning algorithms and conditions, the agent's policy obtained at the end of each run is extracted and analysed (i.e. $24$ policies per algorithm per condition). 

\begin{figure}
    \centering
    \includegraphics{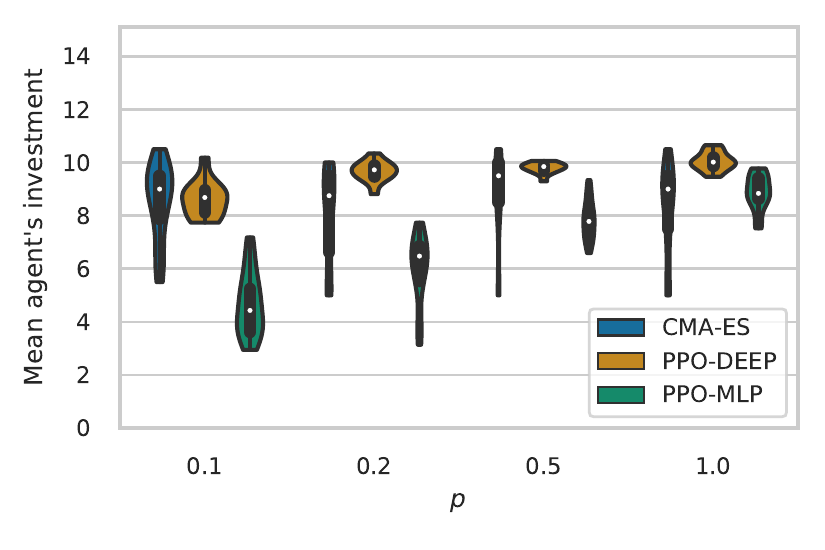}
    \caption{Investment value of the focal agent given by the Investment Module for the best learned policies with CMAES (blue), PPO-DEEP (orange) and PPO-MLP (green) algorithms, for each condition $p$. Each violin graph represents the results of the outcome of the $24$ best policies for a given algorithm and condition after being re-evaluate for $1000$ episodes without learning.}
    \label{fig:investment_comp}
\end{figure}

Figure~\ref{fig:investment_comp} illustrates the outcome of the Investment Module ($x_\bullet$), i.e. the investment value offered by the focal agent when faced with a potential partner. It is obtained by measuring the investment value of the focal agent\footnote{Note that for CMAES the Investment Module follows a deterministic policy (but not the Choice Module). Therefore, it would have been equivalent to take the investment value from the policy parameters in that particular case.} from $1000$ episodes with $p=1.0$ and without learning. Policies learned with CMAES play close to $x_c=10$, which is the optimal play for the payoff function (Section~\ref{ssec:payoff}), whatever the frequency of significant events. As expected, this is different for policies learned with PPO, as the outcome values of the Investment Module are significantly lower when the frequency of significant events decreases ($p<1.0$). 

\begin{figure}
    \centering
    \includegraphics{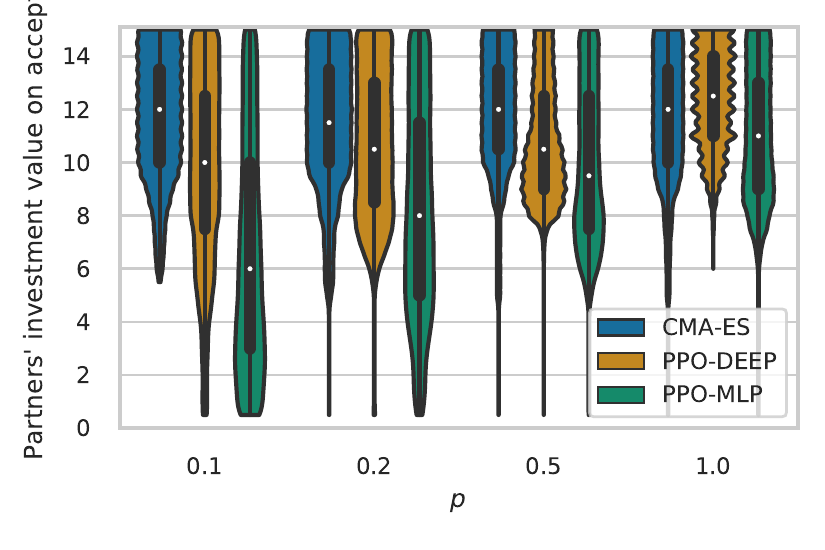}
    \caption{Decision to accept to cooperate taken by the focal agent, when facing a cooperative partner with a particular investment value. Results for CMAES (blue), PPO-DEEP (orange) and PPO-MLP (green) are shown as violin graph. X-axis: algorithms and conditions, Y-axis: partner's investment value for which the focal agent accept to cooperate. 
    }
    \label{fig:accept_comp}
\end{figure}

Figure~\ref{fig:accept_comp} illustrates the investment values played by cooperative partners, when the focal agent accepts to cooperate (whether or not cooperation will actually take place, as it also depends on the partner's acceptance). In other words, it represents how demanding is the focal agent with respects to its partners' intention to invest in cooperation. The probability to accept cooperation is computed for the policies of each run. Each policy is presented with all $31$ possible cooperative partners, $100$ times each, to estimate the focal agent strategy. While CMAES produced consistent policies that follow quasi-identical strategies for all conditions (ie. accepting partners that invest close to the optimal $x_c=10$ or above), this is not the case for PPO policies which are less demanding for lower value of $p$, with many of the policies learned by PPO-MLP with condition $p=0.1$ actually accepting \textit{any} partners). PPO-DEEP policies fare better than PPO-MLP policies, but still worse than policies learned with CMAES when significant events are rarer.
% reminder: partners propose to invest between $0$ and $15$ with step-size: $0.5$. Cf. Section 2.

\begin{figure}
    \centering
    \includegraphics{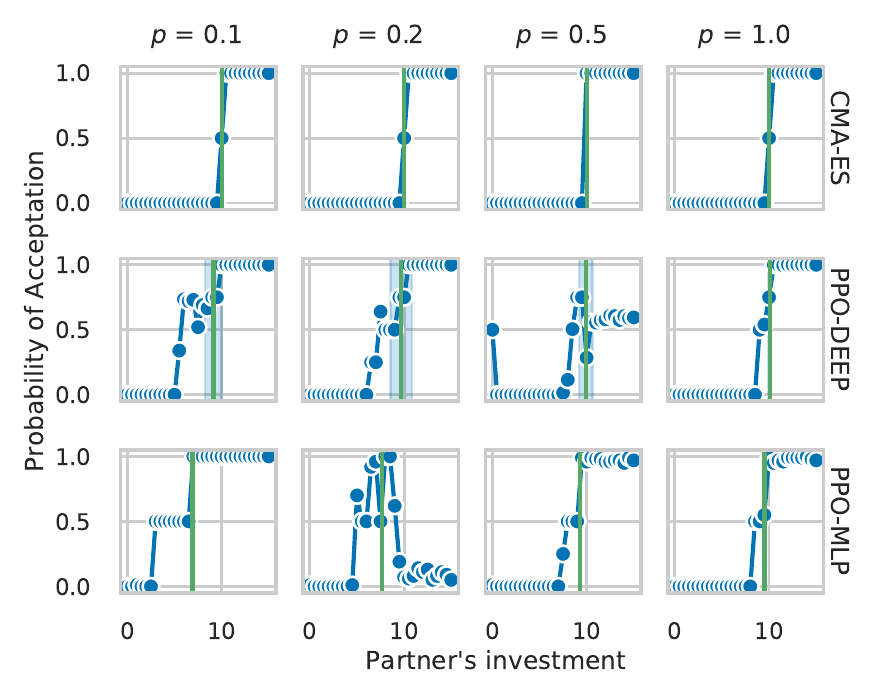}
    \caption{
    Analysis of the Partner Choice module for all conditions (by columns: $p\in \{0.1,0.2,0.5,1.0\}$) and all algorithms (top: CMAES, center: PPO-DEEP, bottom: PPO-MLP). For each setup, only the best policy is shown. Each graph plots the probability to accept cooperation for the focal agent following the best policy (y-axis) depending on its partner's proposed investment (x-axis). Data are computed by presenting each of the $31$ possible cooperative partners to the focal agent for $100$ iterations as policies are stochastic. The green vertical line represent the mean investment of the focal agent. 
    }
    \label{fig:bestpoliciesdetailed}
\end{figure}

Figure~\ref{fig:bestpoliciesdetailed} takes a detailed look at the results shown in Figure~\ref{fig:accept_comp}. It shows the strategy profile for partner choice by the \textit{best} policy obtained with each algorithm in each condition. Focal agents obtained with CMAES follow an efficient and clear-cut strategy: they play the optimal investment value ($x_\bullet= x_c = 10$, green vertical line) and accept partners only when those play a similar or better value ($x_i^+\geq10$, blue line). Policies obtained with PPO-DEEP and PPO-MLP either follow \textit{roughly} the same profile with a more stochastic behaviour (PPO-MLP policies for $p=1.0$ and $0.5$, PPO-DEEP policies for $p=1.0$, $0.2$ and $0.1$) or display a selective strategy, choosing partners only when they play close to the optimal investment value $x_i^+ \approx x_c$. Only PPO-MLP produced policies which are clearly sub-optimal for $p=0.2$ and $p=0.1$, with a mean investment below the optimal investment value $x_\bullet < x_c$.

%%%%%%%%
% NOUVELLE VERSION FIG9
%%%%%%%%

%%%%%%%%%
% FIN NOUVELLE VERSION FIG9
%%%%%%%%%

\section{Concluding Remarks}

In this article, we focused on an on-policy reinforcement learning problem of an autonomous agent that needs to maximize its gain when interacting with other agents, with whom our agent may or may not decide to cooperate. The peculiarity of this problem is to present a (very) small number of significant events during which the agent can obtain only one single positive reward. The challenge is therefore to learn how to best choose a partner, by making a compromise between the chances of finding a better partner, and the cost of an interaction.

We studied the dynamics of two reinforcement learning methods: a gradient policy search algorithm and a direct policy search algorithm with an evolution strategy. Both algorithms succeeded in learning policies that make an optimal use of partner choice when interaction opportunities are frequent. However, the two algorithms differ fundamentally when interaction opportunities are rare. The direct policy search algorithm shows total robustness, while the gradient policy search algorithm collapses, resulting in sub-optimal policies.

The robustness of the direct policy search method can be expected as the sequential and temporal aspects of the task is lost within one evaluation. As long as the evaluation time is long enough to sample the whole population of relevant partners, there is no cost nor change in the algorithm dynamics to deal with a situation where significant events are lost within a longer sequence, but still of the same number. Such independence to action frequency and delayed rewards have actually been observed elsewhere, though for different problems (e.g.: robotic control problem~\citep{Salimans2017}). This is of course different for the gradient policy search method, where increased rarity means that many learning steps will be performed with zero-reward, resulting in poor gradient information most of the time. Not only this slows down learning, even with a similar number of iterations, but it also prevents learning from converging towards a truly optimal partner choice strategy. This remains true even when a large search space is considered, in which over-parametrization in deep neural networks help gradient search~\cite{du2018icml,neyshabur2018arxiv}. 
% it falls short of 
%In policy gradient methods such as PPO, and contrary to direct policy search methods such as CMAES, the very length of the sequence of events actually matters. 

%Previous works from \citet{frank2008} revealed and studied this problem of rare significant events. However, these works were conducted with off-line off-policy reinforcement learning and proposed other methods, such as importance sampling~\citep{ciosek2017}, to address the problem of rare significant events. In our particular case, such methods cannot be applied because of the on-line on-policy nature of the cooperation problem. It would be interesting though, to endow policy gradient methods with similar mechanisms to stand robust in this context.

The broader motivation behind this work is to identify reinforcement learning problems for which evolutionary algorithms as a direct policy search method offer a competitive advantage over gradient policy research methods (cf. also~\citep{igel2003,morse2016,such2017,conti2018,Salimans2017,sigaud2019,Pagliuca2020}). The take-home message that emerges from this paper is that one of these problems occurs when important events are rare, for which direct policy search shows an invariance to rarity. 

As a final remark, it may be tempting to relate the problem of rare significant events with that of sparse rewards, which has gain a lot of attention recently~\citep{konidaris2006, jaderberg2017, riedmiller2018}. However, they differ fundamentally as significant events may be rare, but \textit{eventually} occur. This is not the case with sparse rewards, which occurrences are conditioned by the policy itself (e.g. a robotic arm must be within the length of a target to trigger a reward) and may \textit{never} be obtained. We also argue that problems where significant events are rare rather than sparse may be more numerous than expected: a complex environment offers multiple learning opportunities, as long as one is able to seize them as they arise.

\begin{acks}
This work is funded by the Agence Nationale pour la Recherche under Grant No ANR-18-CE33-0006. We would like to thank Yann Chevaleyre, Olivier Sigaud and Mathieu Seurin for feedbacks and comments.
\end{acks}

% CONTENT
% Detail analysis of the agents’ reward
% Re-evaluation performance statistical score
% Timing of the algorithm (short answer: CMAES is much faster)
% Influence of the discount factor
% Influence of the absence of critic

\bibliographystyle{ACM-Reference-Format}
\bibliography{allref.bib}

% ANONYMOUS 
\clearpage
%\appendix
%\section{Supplementary materials}\label{supmat}
%\input{supmat}
% ANONYMOUS 

% \section*{Supplementary materials}\label{supmat}
% Document attached to submission.

\end{document}

\maketitle

\appendix
\section{Supplementary materials}

\subsection{Detail analysis of the agents' reward}

\begin{figure}[H]
    \centering
    \includegraphics{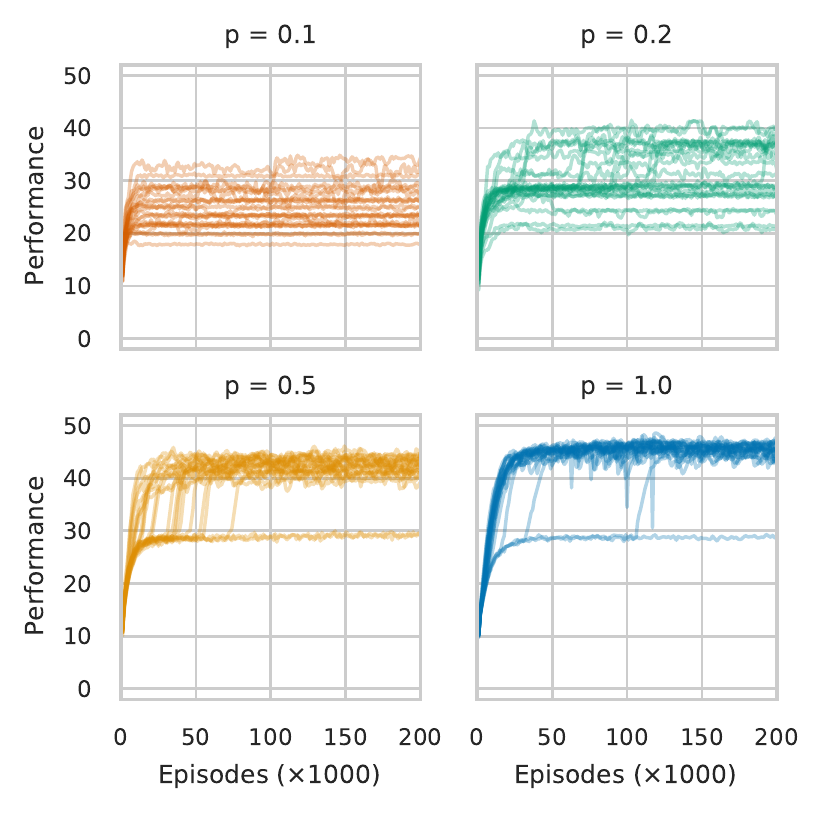}
    \caption{Split view of the different performances with PPO-MLP. For $p = 1.0$, in most of the simulations, the agent reaches a performance close to the optimal performance at the end of its learning. In a few simulations, the agent gets stuck at a plateau value. If the agent overcome this plateau value, it reaches quickly the close-to-optimal performance. For $p = 0.5$, in most of the simulation the agent's performance gets close to the optimal reward value. The plateau is reached more often than in the $p = 1.0$ condition. The plateau's equilibrium is even stronger for $p = 0.2$ and $p = 0.1$. In almost no simulation to no simulation at all the agent's performance escape from the plateau equilibrium to reach the close-to-optimum equilibrium.}
    \label{fig:pporewarddetail}
\end{figure}

We plot the performance per run of each conditions in Fig.~\ref{fig:pporewarddetail}. As $p$ gets lower, the performance reached by a run gets slightly lower. Furthermore, for $p \in \{1, 0.5\}$, agent's reward converges to two different values: one optimal equilibrium above 40 and a second sub-optimal equilibrium at around 30. There are two equilibria that the algorithm reaches. The transition from the sub-optimal equilibrium to the optimal equilibrium is quick, but occurs less and less often as $p$ gets smaller.

The detailed analysis of the performances through the episodes of the CMAES condition shows the same pattern as the PPO-MLP conditions for $p=1.0$. For smaller values of $p$, CMA-ES agents gets still close to the optimal performance, whereas PPO-MLP agents' performances plunger.

\begin{figure}[H]
    \centering
    \includegraphics{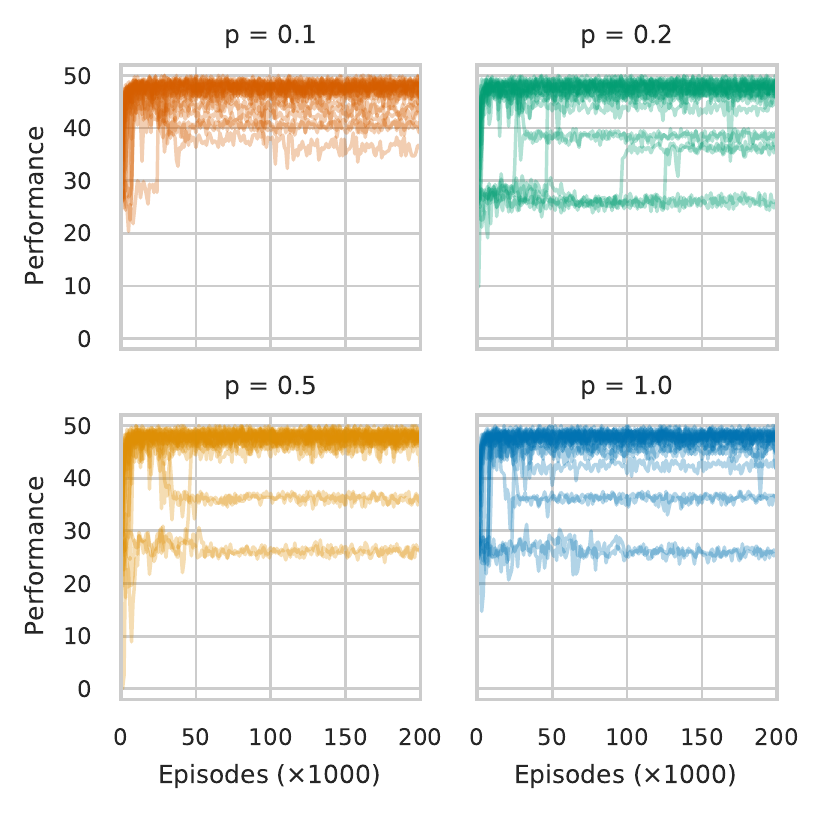}
    \caption{Split view of the different performances of CMA-ES for different values of $p$. Regardless of the value $p$, in most of the simulations the agent gets a performance close to the optimal performance at the end of the learning. Like the PPO-MLP condition, some simulations reach a sub-optimal equilibrium around 30. Few of them manage to get out of this equilibrium.}
    \label{fig:cmarewarddetail}
\end{figure}

The detailed analysis of the performance through the episodes of the PPO-DEEP condition shows the same pattern as the PPO-MLP conditions. Though, the performance loss as $p$ gets lower is far less strong.

\begin{figure}[H]
    \centering
    \includegraphics{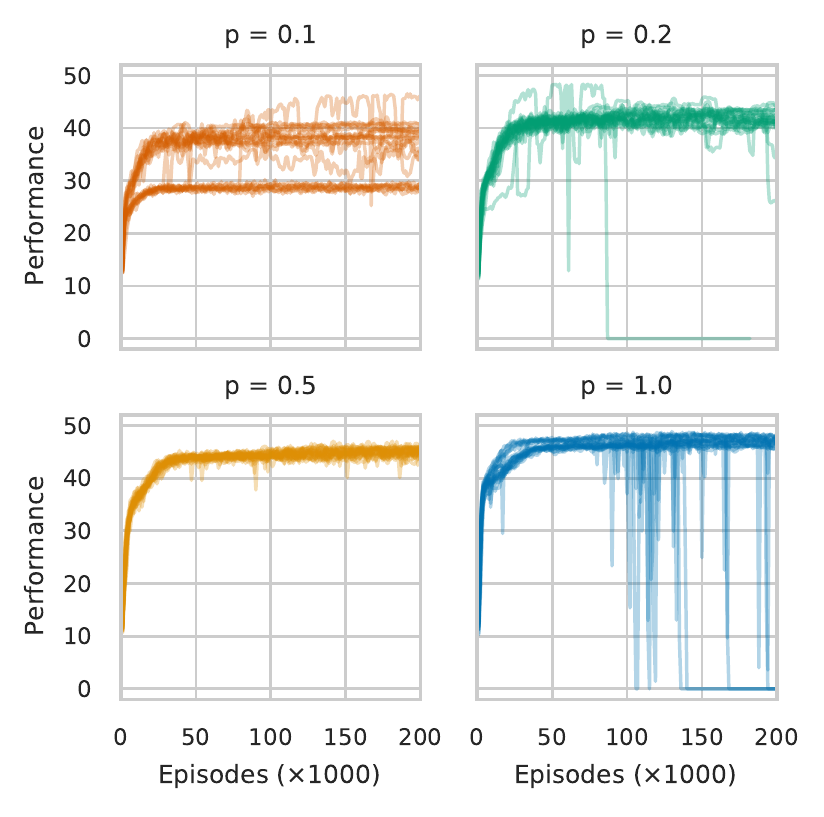}
    \caption{Split view of the different performances of PPO-DEEP for different values of $p$. Performance varies with $p$, the higher $p$, the better the performance. The loss in performance as $p$ gets lower is less strong than with PPO-MLP. Performance is highly unstable after 100,000 generations for $p = 1.0$.}
    \label{fig:ppodeeprewarddetail}
\end{figure}

\subsection{Re-evaluation performance statistical score}

We perform two-tailed Mann-Whitney's U-tests to compare the distributions of the performance for the PPO-MLP, PPO-DEEP and CMA-ES agent for each probability $p$ of meeting a $x^+$ partner. The table of the median performance of each learning algorithm for each $p$ is reported in table~\ref{tab:median_perf} and the U-statistics and p-values of the tests are reported in table~\ref{tab:stattable}.

\begin{table}[H]
    \centering
\begin{tabular}{llrrrr}
\toprule
     &         &  Median &   MAD &  Mean &   Std \\
$p$ & Algorithm &         &       &       &       \\
\midrule
0.10 & CMA-ES &   47.72 &  2.45 & 46.19 &  3.21 \\
     & PPO-DEEP &   37.83 &  4.43 & 35.73 &  5.16 \\
     & PPO-MLP &   24.04 &  3.43 & 24.92 &  4.23 \\
0.20 & CMA-ES &   47.56 &  5.32 & 44.03 &  6.98 \\
     & PPO-DEEP &   41.18 &  1.90 & 40.76 &  3.57 \\
     & PPO-MLP &   32.44 &  5.20 & 31.60 &  5.93 \\
0.50 & CMA-ES &   48.00 &  4.59 & 45.17 &  6.67 \\
     & PPO-DEEP &   45.23 &  0.70 & 45.06 &  0.89 \\
     & PPO-MLP &   42.52 &  2.62 & 41.49 &  4.17 \\
1.00 & CMA-ES &   47.64 &  4.72 & 44.42 &  6.62 \\
     & PPO-DEEP &   46.99 & 13.04 & 39.12 & 17.89 \\
     & PPO-MLP &   45.58 &  1.86 & 44.78 &  3.62 \\
\bottomrule
\end{tabular}

    \caption{Median of the re-evaluations, 24 runs per condition}
    \label{tab:median_perf}
\end{table}

\begin{table}[H]
    \centering
\begin{tabular}{llrr}
\toprule
    &                    &          $\mathit{p-value}$ &  U-statistic \\
$p$ & test &               &              \\
\midrule
0.1 & PPO-MLP vs PPO-DEEP & 1.2e-07 &           31 \\
    & PPO-MLP vs CMA-ES & 3.1e-09 &            0 \\
    & PPO-DEEP vs CMA-ES & 3.9e-08 &           21 \\
0.2 & PPO-MLP vs PPO-DEEP & 2.5e-07 &           33 \\
    & PPO-MLP vs CMA-ES &   3e-06 &           61 \\
    & PPO-DEEP vs CMA-ES &  0.0023 &          132 \\
0.5 & PPO-MLP vs PPO-DEEP & 6.4e-07 &           46 \\
    & PPO-MLP vs CMA-ES & 5.5e-05 &           92 \\
    & PPO-DEEP vs CMA-ES & 9.3e-05 &           98 \\
1.0 & PPO-MLP vs PPO-DEEP &   0.018 &          173 \\
    & PPO-MLP vs CMA-ES &   0.019 &          174 \\
    & PPO-DEEP vs CMA-ES &    0.12 &          212 \\
\bottomrule
\end{tabular}

    \caption{Statistical results of the two-tailed Mann-Whitney U-test comparing the performance of the agents using PPO-MLP, PPO-DEEP and CMAES in the re-evaluation setup. $n=24$ for each condition and algorithm.}
    \label{tab:stattable}
\end{table}

\subsection{Timing}

We measure the execution time (wall time) for both PPO-MLP and CMAES on a single CPU. To do so, we take the best agent out of the 24 simulations for each algorithm and each condition, and restart the learning from its state for 15 minutes. We then divide the total time taken by the algorithm by the number of episode time step the algorithm fulfilled. 

CMA-ES overhead is constant, as CMA-ES updates are more frequent per episode time steps as $p$ increases. Indeed, the shorter the episodes are, the more often CMA-ES updates triggers. PPO-MLP and PPO-DEEP updates are constant with respect to the episode length, as they always update when 4000 episode time steps are completed. The PPO-MLP execution time compare to the CMA-ES execution time is explain by the RLlib overhead as well as its learning computational cost. The PPO-DEEP execution time is greater than PPO-MLP execution time. It is explained both by the larger network that requires a more computationally intensive evaluation, as well as the gradient descent which involve massively more weights.

\begin{figure}[H]
    \centering
    \includegraphics{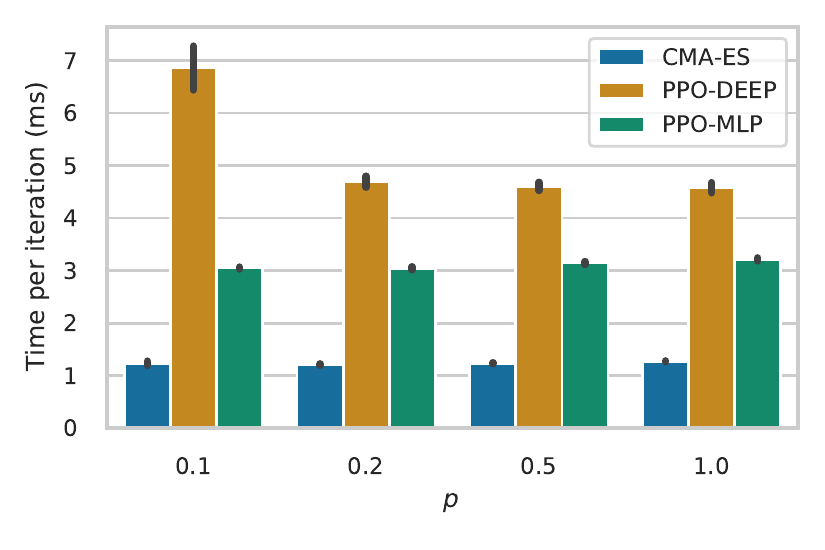}
    \caption{Average time per episode time steps (Learning included) for CMAES, PPO-DEEP and PPO-MLP for the different conditions $p$.}
    \label{fig:time_per_it}
\end{figure}

The total time divided by the number of iteration is reported in Figure~\ref{fig:time_per_it} as well as in Table~\ref{tab:computation_time}. Speed ratio are shown in Table~\ref{tab:speedratio}

\begin{table}[H]
    \centering
    \begin{tabular}{llr}
\toprule
        &      & Time per step (ms) \\
Algorithm & $p$ &   Mean $\pm$ std\\
\midrule
CMA-ES & 0.10 &          $1.24 \pm 0.04$ \\
        & 0.20 &          $1.21 \pm 0.01$ \\
        & 0.50 &          $1.24 \pm 0.01$ \\
        & 1.00 &          $1.27 \pm 0.01$ \\
PPO-DEEP & 0.10 &          $6.86 \pm 0.43$ \\
        & 0.20 &          $4.69 \pm 0.11$ \\
        & 0.50 &          $4.60 \pm 0.08$ \\
        & 1.00 &          $4.58 \pm 0.10$ \\
PPO-MLP & 0.10 &          $3.05 \pm 0.02$ \\
        & 0.20 &          $3.04 \pm 0.03$ \\
        & 0.50 &          $3.14 \pm 0.03$ \\
        & 1.00 &          $3.21 \pm 0.03$ \\
\bottomrule
\end{tabular}

\begin{table}[H]
    \centering
    \begin{tabular}{lrrr}
\toprule
Speed ratio &  CMA-ES &  PPO-DEEP &  PPO-MLP \\
col/row &         &           &          \\
\midrule
CMA-ES   &    1.00 &      4.18 &     2.51 \\
PPO-DEEP &    0.24 &      1.00 &     0.60 \\
PPO-MLP  &    0.40 &      1.66 &     1.00 \\
\bottomrule
\end{tabular}

    \caption{Speed ratio between algorithms}
    \label{tab:speedratio}
\end{table}

    \caption{Table of computational wall time per environmental time step for CMA-ES, PPO-DEEP and PPO-MLP with one single core. The average time per environmental step includes the time needed by the learning algorithm to update the policy.}
    \label{tab:computation_time}
\end{table}

\subsection{Influence of the discount factor}

\begin{figure}[H]
    \centering
    \includegraphics[width=\linewidth]{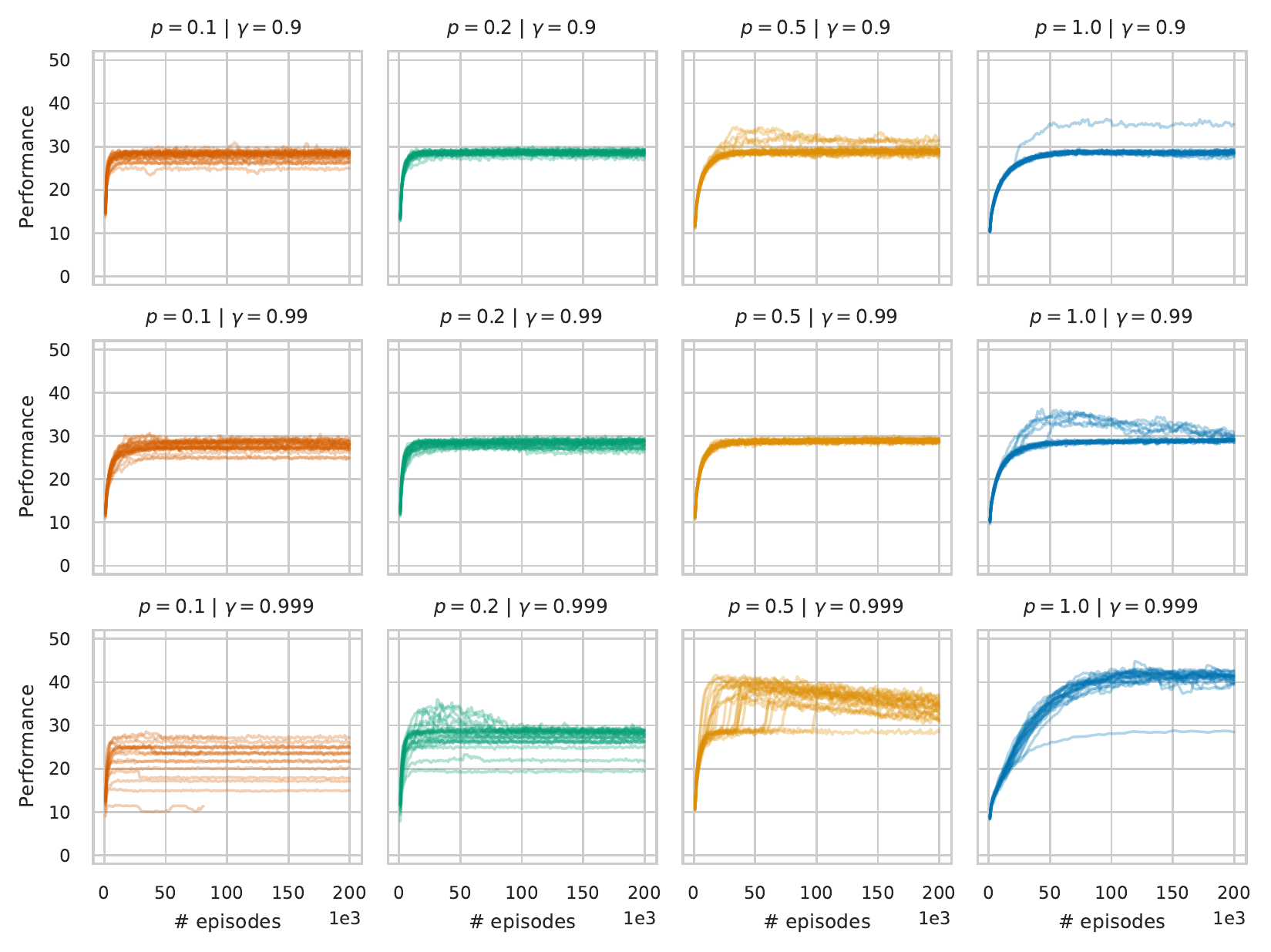}
    \caption{Performance of the PPO-MLP agent for different values of discount factor $\gamma \in \{0.9, 0.99, 0.999\}$. Low $\gamma$ has a negative impact on performance compared to $\gamma = 1$. The best results are obtained for $\gamma = 1$.}
    \label{fig:gammareward}
\end{figure}

We tested different value of the discount factor $\gamma$ for PPO-MLP. These results are reported in Figure~\ref{fig:gammareward}. Using a $\gamma < 1$ has a detrimental impact on the agent's performance regardless of the value of $p$. The agent is always stuck in the suboptimal plateau of a performance of 25. With $\gamma = 0.999$ and $p=1.0$, the agent gets a performance close to the performance of the PPO-MLP agents with $\gamma = 1$.

%\subsection{Influence of the size of the network}

%We wanted to test the impact of the network size for the choice module. Indeed, having more dimension tot explore for PPO prevent the algorithm from getting stuck in local optima, and therefore could help the performance of the algorithm. Choosing a network size of two layers of 256 neurons had no impact or deleterious impact compared to the smaller network, as shown in Fig~\ref{fig:bignet_rewards}.

%\begin{figure}[H]
%    \centering
%    \includegraphics{media/results/appendix/bignet_reward.pdf}
%    \caption{Performance of the PPO agent with a bigger choice  network with two hidden layers of 256 neurons. The impact of the bigger network is negligeable.}
%    \label{fig:bignet_rewards}
%\end{figure}

\subsection{Influence of the absence of critic}

The absence of a critic does not have any positive influence on the performance of the PPO-MLP agent. The performance is much nosier and below the performance of the PPO-MLP agent with a critic and GAE.

\begin{figure}[H]
    \centering
    \includegraphics{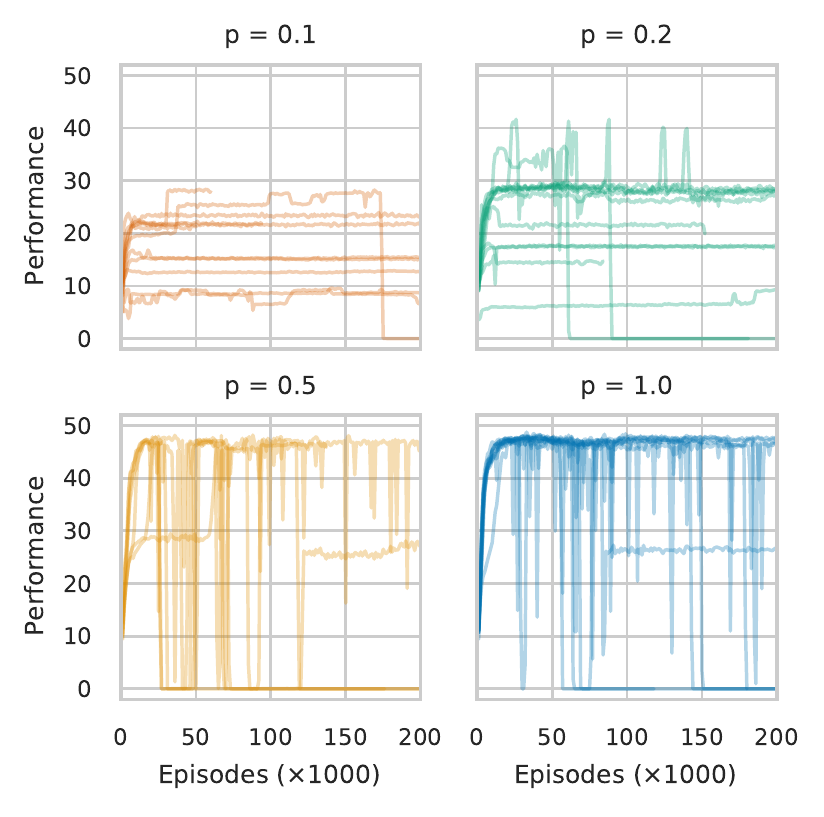}
    \caption{The absence of critic does not improve PPO performance.}
    \label{fig:nocritic_mean_reward}
\end{figure}

%%%%%%%%%%%%%%%%%%%%%
% NOUVELLES FIGURES %
%%%%%%%%%%%%%%%%%%%%%

\subsection{PPO Loss Function}

%%%% Loss function PPO
Our PPO implementation optimize $\theta$ by maximising the function $LL^{CLIP+KLPEN}(\theta)$ given in Equation~\ref{eq:ppoloss}.

\begin{equation}
\begin{split}
L^{CLIP+KLPEN}(\theta)=&\hat{\mathbb{E}}_{t}[\min \left(r_{t}(\theta) \hat{A}_{t}, \operatorname{clip}\left(r_{t}(\theta), 1-\epsilon, 1+\epsilon\right) \hat{A}_{t}\right) \\
&-\beta \mathrm{KL}\left[\pi_{\theta_{\mathrm{old}}}\left(\cdot \mid s_{t}\right), \pi_{\theta}\left(\cdot \mid s_{t} \right) \right] ] \\
\end{split} \label{eq:ppoloss}
\end{equation}

with $r_{t}(\theta)=\frac{\pi_{\theta}\left(a_{t} \mid s_{t}\right)}{\pi_{\theta_{\text {old }}}\left(a_{t} \mid s_{t}\right)}$, $\hat{A}_t$ the estimator of the advantage at time step $t$. $\beta$ varies according to the KL target $d_{targ}$ (here 0.01). $\beta = 0.2$ at the beginning of the simulation and adjusts at each time step according to the following rule, be $d=\mathbb{E}_{t}\left[\operatorname{KL}\left[\pi_{\theta_{\text {old }}}\left(\cdot \mid s_{t}\right), \pi_{\theta}\left(\cdot \mid s_{t}\right)\right]\right]$ :

\begin{equation}
\left\{
    \begin{array}{ll}
        \beta \leftarrow \beta \times 1.5 & \mbox{if $d > 2\times d_{targ}$} \\
        
        \beta \leftarrow \beta / 2 & \mbox{if $d < d_{targ}/2$}
    \end{array}\right.
\end{equation}

All the PPO-MLP and PPO-DEEP hyperparameters are available in Table~\ref{tab:ppo_parameters_full}.

\begin{table}[H]
    \centering
    \begin{tabular}{cc}
        \toprule
        \textbf{Parameters} & \textbf{Values} \\
        \midrule
        Learning rate PPO-MLP & $0.005$\\
        Learning rate PPO-DEEP & $0.001$\\

        Optimiser Algorithm & SGD \\
        Number of optimisation epochs & $10$\\
        Minibatch size & $128$ \\
        Batch size & $4000$ \\
        Kullback-Leibler coefficient $\beta$ & 0.2 \\
        Kullback-Leibler target $d_{targ}$ & 0.01 \\ 
        Discount factor $\gamma$ & 1.0 \\
        Search space PPO-MLP ($\theta_{MLP}$) & $\mathbb{R}^{33}$\\
        Search space PPO-DEEP ($\theta_{DEEP}$) & $\mathbb{R}^{133894}$\\
        \bottomrule
    \end{tabular}
    \caption{All parameters for the PPO algorithm}
    \label{tab:ppo_parameters_full}
\end{table}